\documentclass[journal]{IEEEtran}
\usepackage{graphicx}
\usepackage{amssymb}
\usepackage{color}
\usepackage{graphicx}
\usepackage{caption}
\usepackage{subcaption}
\usepackage[utf8]{inputenc}

\usepackage{amssymb}
\usepackage{color}

\begin{document}
\title{Automated Multiscale 3D Feature Learning\\for Vessels Segmentation in Thorax CT Images}

%
%

\author{
\IEEEauthorblockN{Tomasz Konopczyński$^{1,2,3}$, Thorben Kröger$^{3}$, Lei Zheng$^{1}$, Christoph S. Garbe$^{2}$, Jürgen Hesser$^{1,2}$\bigskip}

\IEEEauthorblockA{
$^1$ Experimental Radiation Oncology, Dept. of Radiation Oncology,\\
University Medical Center Mannheim, Heidelberg University, Heidelberg, Germany\\
$^2$ Interdisciplinary Center for Scientific Computing (IWR), Heidelberg University, Heidelberg, Germany\\
$^3$ Volume Graphics GmbH, Heidelberg, Germany\\
}
}

\maketitle
\pagestyle{empty}
\thispagestyle{empty}

\begin{abstract}
We address the vessel segmentation problem by building upon the multiscale feature learning method of Kiros et al., which achieves the current top score in the VESSEL12 MICCAI challenge. Following their idea of feature learning instead of hand-crafted filters, we have extended the method to learn 3D features. The features are learned in an unsupervised manner in a multi-scale scheme using dictionary learning via least angle regression. The 3D feature kernels are further convolved with the input volumes in order to create feature maps. Those maps are used to train a supervised classifier with the annotated voxels. In order to process the 3D data with a large number
of filters a parallel implementation has been developed. The algorithm has been applied on the example scans and annotations provided
by the VESSEL12 challenge.
We have compared our setup with Kiros et al. by running their implementation. Our current results show an improvement in accuracy over the slice wise method from 96.66$\pm$1.10\% to 97.24$\pm$0.90\%.

\end{abstract}

\section{Introduction}
%
%
%
%
\IEEEPARstart{T}{he} challenge in vessel segmentation from 3D CT scans is to identify thin vessels at low resolution and the ability to distinguish them from other, similar looking structures.
Most of the top-scoring methods in the MICCAI VESSEL12 challenge [1] are based on cost functions derived from the eigenvalues of the Hessian on the image.
Such methods usually differ in the selection of scales [2] and interpretation of eigenvalues [3,4].
Others [5] focus on designing new application-specific kernels.

Feature learning methods via sparse coding have recently become popular for image classification and natural language processing [7].
Instead of using handcrafted filters like in the Hessian-based approaches, the idea is to automatically compute a set of convolution filters particularly tuned to the dataset at hand.
The method which has achieved the current highest score in the VESSEL12 challenge [1,6] uses dictionary learning to obtain convolution filters in an unsupervised manner.
All filter responses form a set of feature maps, which are then used to train a supervised classifier for the vessel/non-vessel voxels.
Whereas Kiros et al. [6] have applied the dictionary learning for 2D patches after designating a preferred slicing orientation, our method extends it to a true 3D approach by using 3D patches.
Learning 3D features instead of slice-wise 2D features should increase the accuracy of the classifier at the expense of computation time and memory consumption.


\section{Method}

The method is divided into two main parts. Dictionary learning for feature learning and Classifier learning for voxel-wise classification of vessel or non-vessel.
In order to capture vessels at different scales Gaussian pyramids from the input volumes are generated.
We use the multi-scale representation during both feature learning (i.e. our patches are sampled from original and scaled volumes at different scales) and classifier learning (i.e. we extract feature maps at many scales).

\subsection{Dictionary Learning}

\begin{figure}[!t]
    \centering
    \includegraphics[width=3.5in]{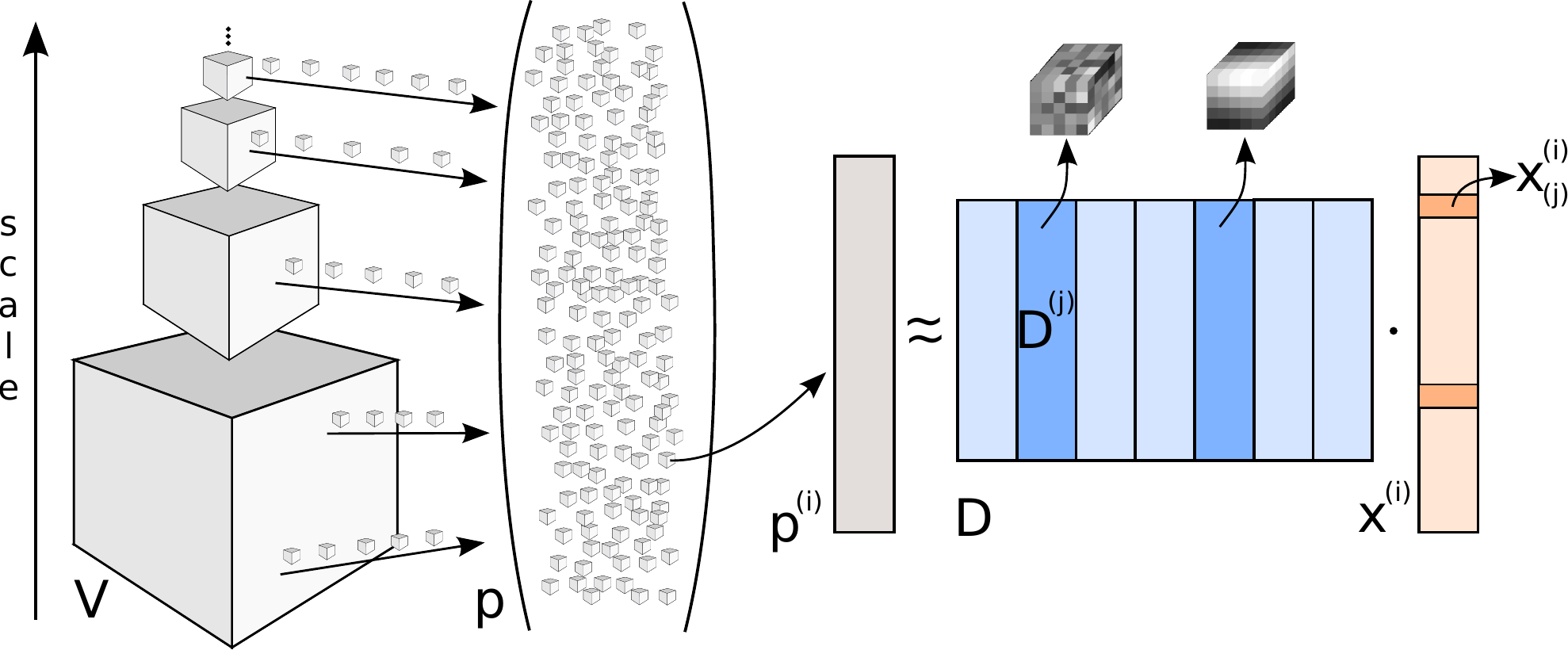}
    \caption{\textbf{Dictionary learning overview.} From a given volume $V$ a random batch of patches $p$ is extracted. Patches have the same dimensionality as $V$, but are extracted from different scales of the volume. Dictionary elements $D^{(j)}$ in a dictionary $D$ are learnt in a way to make it possible to reconstruct each patch $p^{(i)}$ by a \emph{sparse} linear combination of the elements. In other words $p^{(i)}$ is approximated by a product of the dictionary and a sparse vector $x^{(i)}$ . }\label{fig:dictionary}
\end{figure}

Feature generation via Dictionary Learning is an unsupervised problem, where from a number of patches the algorithm learns a set of elements that allow for an optimal representation.
First we create a set of Gaussian pyramids by convolving input volumes with 3D Gaussian kernels and subsampling them.
From the resulting volume data we randomly select a batch of 3D patches $p^{(i)} \in \mathbb{R}^{n}$ (where $n$ is the number of voxels of the patch) which we use as an input to the sparse coding algorithm for learning a dictionary $D \in \mathbb{R}^{n \times d}$ of $d$ elements, where each column $D^{(j)}$ is one element.
We train the dictionary by minimizing the LASSO problem with L1-penalization to ensure the sparsity of the vector $x^{(i)}$ regularized by the parameter $\lambda$. That is, we optimize 
\begin{equation}
    \min_{D, x^{(i)}} \sum_{i} {||Dx^{(i)} - p^{(i)} ||}_{2}^{2} + \lambda||x^{(i)}||_{1}
    \label{eqn:wzor11}
\end{equation}
\begin{center}
subject to $||D^{(j)}||_{2}^{2} = 1,\forall j $
\end{center}
over the sparse codes $x^{(i)}$ and the dictionary, $D$.
Figure 1 sketches the dictionary learning step.

\subsection{Classifier Learning}

To classify each voxel as vessel or non-vessel, we apply supervised learning, making use of manual voxel-wise annotations.
As features, we take the convolution response of each element $d$ from the previously learned dictionary $D$ with the annotated volume images at each scale $s$.
Thus, each voxel of a volume contains $s \times d$ predictors. 
As the number of features is higher than the number of labels in our case, we opt for a linear logistic regression classifier. Figure 2 presents a graphical representation of the step.
The hyper parameters of the classifier are tuned using 10-fold cross validation.

\begin{figure}[!t]
    \centering
    \includegraphics[width=3.5in]{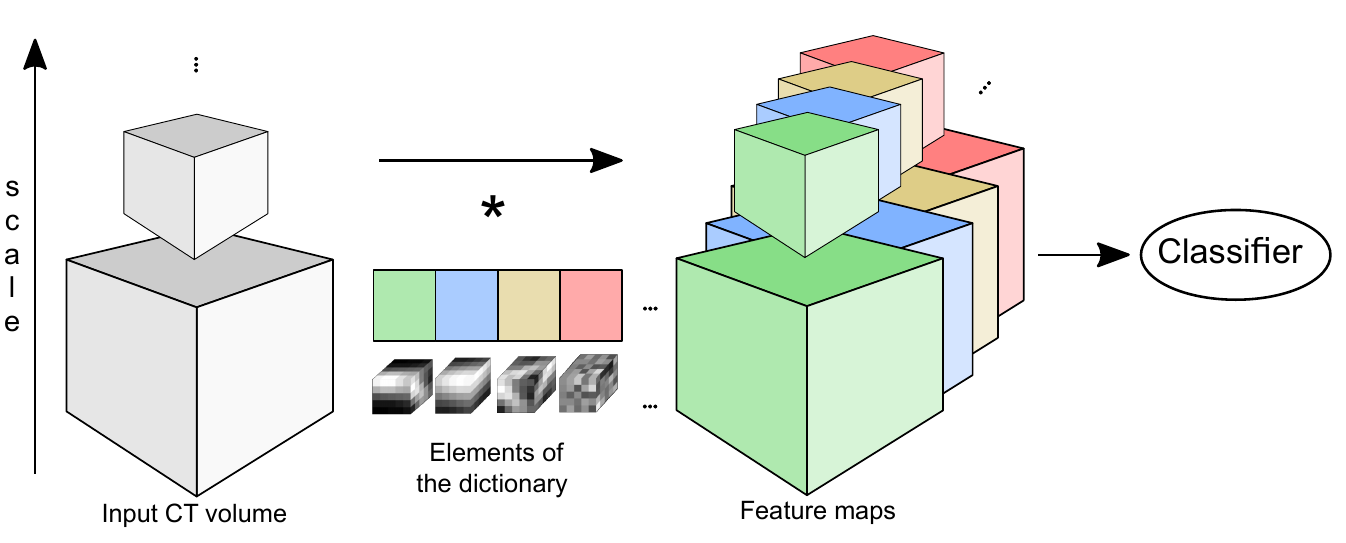}
    \caption{\textbf{Feature extraction and classifier learning.} The annotated volumes are convolved with the elements of the dictionary producing feature maps at a number of scales. That is one feature volume is created per element and scale. The classifier is learnt on the annotated voxels only. The trained classifier can be evaluated on the entire volume.}\label{fig:dictionary}
\end{figure}

\section{Comparison with the 2D approach}

We have evaluated our method on the data set provided by VESSEL12 [1].
The data contains 3 annotated volumes with in total 882 annotated voxels (vessel/non-vessel) and additional 20 volumes without annotations.
Each volume is of dimension 512$\times$512$\times$512.
For each volume, the lung region is denoted by a corresponding mask.

\begin{figure}[!h]
    \centering
    \begin{subfigure}[b]{1in}
        \includegraphics[width=\textwidth]{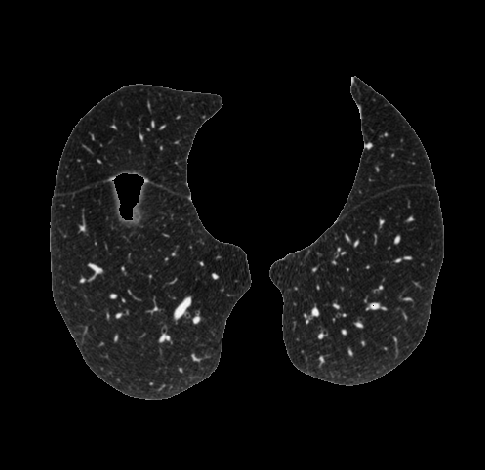}
        \caption{}
        \label{fig:gull}
    \end{subfigure}
    ~ 
    \begin{subfigure}[b]{1in}
        \includegraphics[width=\textwidth]{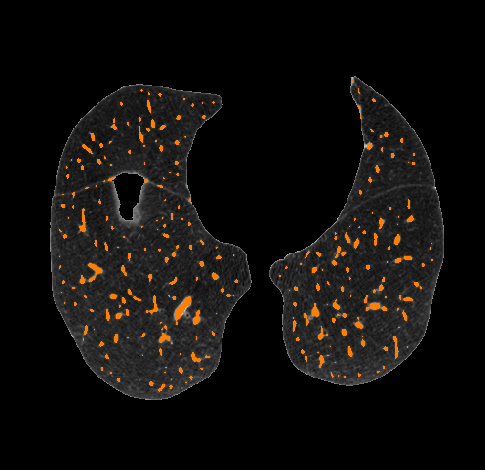}
        \caption{}
        \label{fig:gull}
    \end{subfigure}
    ~ 
    \begin{subfigure}[b]{1in}
        \includegraphics[width=\textwidth]{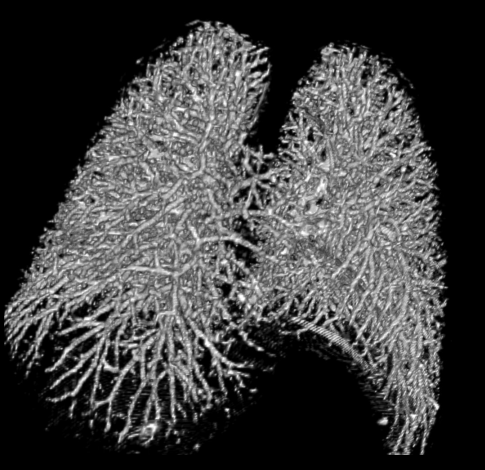}
        \caption{}
        \label{fig:tiger}
    \end{subfigure}
    \caption{\textbf{Result of classification based on learned features.}\\ (a) An example slice from the dataset (with lungmask applied), (b)~regions classified as vessels by our method (in orange), (c)~the~3D visualisation of the regions classified as vessels.}\label{fig:animals}
\end{figure}

For this work we use a mini batch implementation with least angle regression as a minimizer of the dictionary learning on the 20 given volumes without annotations to train the dictionary $D$ with 512 elements $D^{(j)}$ and $\lambda=1$.
We extract 2,450,000 patches from all available volume images at different scales of the  Gaussian Pyramid and normalize them by subtracting the mean.
The calculated elements of the dictionary are later used to create feature maps for the 3 annotated volumes from which the 882 annotated voxels are extracted.
We divide those voxels at random into a training data set of 657 voxels with which we train the classifier and test data set of 225 voxels with which we evaluate the performance.
Annotated voxels are assigned to the training and the test data sets at random.
This step is repeated 1000 times in order to average the performance.
For the classification learning we have achieved best results by using Newton's Method with L2 and number of scales $s=2$.

Our current results for the above setup reached an accuracy of 97.24$\pm$0.90\%
compared to 96.66$\pm$1.10\% by the slice-wise approach.
For the comparison we have used the code and setup provided by Kiros et al. [6].
Figure 3 presents one visual output of our algorithm, and Table 1 shows the comparison between the setups.


\section{Conclusions and Future Work}

We have shown that the 3D approach yields competitive results.
We found that using more than 2 scales did not increase the accuracy.
One reason may be that the number of features gets too large compared to the number of annotated examples.
In the future we plan to gather more training data and use a feature selection method.
A significantly higher number of examples should allow us to use more scales and a more flexible model.

Though we have used the source code and setup directly provided by Kiros et al., they have reported to use an additional step for the VESSEL12 challenge. Concretely they have used an additional layer of depth, to learn additional features from the derived feature maps. This area is worth exploring in future research.


\begin{table}[!t]
\renewcommand{\arraystretch}{1.3}
\caption{Comparison of the results with corresponding parameters.}
\label{table_example}
\centering
\begin{tabular}{c|c|c}
\bfseries   & \bfseries Kiros et al. & \bfseries Our method \\
\hline\hline
Algorithm for $D$ training & OMP-1  & minibatch LASSO+LAR\\
\hline
Algorithm for logit regression & LBFGS L2  & Newton's L2\\
\hline
Dim of patches and elements & 5$\times$5 & 5$\times$5$\times$5\\
\hline
Number of patches $p^{(i)}$ & 100,000 & 2,500,000\\
\hline
Number of elements $D^{(j)}$ & 32  & 512\\
\hline
Number of scales $s$ & 6 & 2\\
\hline
Number of features & 192 & 1024\\
\hline\hline
Accuracy & \textbf{ 96.66$\pm$1.10\%} & \bfseries 97.24$\pm$0.90\% \\
\end{tabular}
\end{table}

\end{document}